\documentclass{article}
\usepackage[english]{babel}

\usepackage[letterpaper,top=2cm,bottom=2cm,left=3cm,right=3cm,marginparwidth=1.75cm]{geometry}

\usepackage{amsmath}
\usepackage{graphicx}
\usepackage[colorlinks=true, allcolors=blue]{hyperref}
\usepackage{booktabs}
\usepackage{xspace}
\usepackage{color}
\usepackage{soul}
\usepackage{colortbl}
\usepackage{multirow}
\usepackage[table,xcdraw]{xcolor}
\usepackage{float}
\usepackage{hyphenat}
\usepackage{parskip}

\title{Detecting Autism Spectrum Disorders with Machine Learning Models Using Speech Transcripts \vspace{.75em}}
\author{
  Vikram Ramesh\\ WWP High School North \\
  \texttt{22VR0591@wwprsd.org}
  \and
  Rida Assaf\\ University of Chicago \\
  \texttt{rida@cs.uchicago.edu}
}

\usepackage{titling}
\begin{document}\twocolumn
\date{}\predate{}\postdate{} 
\maketitle
\begin{sloppypar}
\begin{abstract}
Autism spectrum disorder (ASD) can be defined as a neurodevelopmental disorder that affects how children interact, communicate and socialize with others. This disorder can occur in a broad spectrum of symptoms, with varying effects and severity. While there is no permanent cure for ASD, early detection and proactive treatment can substantially improve the lives of many children. Current methods to accurately diagnose ASD are invasive, time-consuming, and tedious. They can also be subjective perspectives of a number of clinicians involved, including pediatricians, speech pathologists, psychologists, and psychiatrists. New technologies are rapidly emerging that include machine learning models using speech, computer  vision from facial, retinal, and brain MRI images of patients to accurately and timely detect this disorder. Our research focuses on computational linguistics and machine learning using speech data from TalkBank, the world’s largest spoken language database. We used data of both ASD and Typical Development (TD) in children from TalkBank to develop machine learning models to accurately predict ASD. More than 50 features were used from specifically two datasets in TalkBank to run our experiments using five different classifiers. Logistic Regression and Random Forest models were found to be the most effective for each of these two main datasets, with an accuracy of 0.75. These experiments confirm that while significant opportunities exist for improving the accuracy, machine learning models can reliably predict ASD status in children for effective diagnosis.
\end{abstract} \\


\section{Introduction}

The Centers for Disease Control (CDC) considers Autism Spectrum Disorder (ASD) as a developmental disability that begins around 3 years of age and can last a lifetime in a person. Symptoms may begin in children as young as 12 months but may also begin at a later stage (24 months or even more). Children with ASD typically struggle with their ability to interact and communicate in social settings. They display restricted behaviors or interests, repeat themselves and exhibit other characteristics like delayed language, cognitive, or learning skills, hyperactive and eccentric activities. About 1\% of the world population [1] and an estimated 5.4M people in the US have ASD [2], affecting 1 in 54 children. Roughly 25\% of children with ASD go undiagnosed for various reasons, including painstaking and expensive methods currently used.

ASD can be detected by early surveillance and developmental monitoring, but the process is long, elaborate, and tedious [3] with multiple screening stages at various ages of the child, with parents completing a series of checklists and questionnaires and children’s development monitored regularly. This is followed by more intensive questioning of parents and screening to conclusively diagnose and develop treatment plans. Unfortunately, early symptoms for ADHD are also similar to the symptoms of ASD, and many doctors misdiagnose the condition, causing anxiety and frustration for parents and deterioration of ASD in children.

Research shows that early detection and intervention for ASD can lead to long-term positive effects and help children reach their fullest potential [4]. For milder forms of ASD like Asperger’s syndrome [5], early diagnosis and treatment can help children develop healthy social and communication skills. It will also help them to learn new skills to become more independent and successful. However, accurate diagnosis is often delayed because ASD characteristics may not emerge until the disorder is well established.

\begin{table*}[t]
\centering
\begin{tabular}{|c|c|c|c|c|c|c|c|}
\hline
\rowcolor[HTML]{E7E6E6} 
\textbf{TalkBank} &
  \textbf{Group} &
  \textbf{Sex} &
  \textbf{Participants} &
  \textbf{Age{[}1{]}} &
  \textbf{Total Words{[}2{]}} &
  \textbf{Mean{[}3{]}} &
  \textbf{Median{[}4{]}} \\ \hline
                                                &                                              & M & 13 & 40 & 518 & 3.4 & 2.8  \\ \cline{3-8} 
                                                & \multirow{-2}{*}{TD}                         & F & 3  & 45 & 383 & 2.6 & 1.75 \\ \cline{2-8} 
                                                &                                              & M & 14 & 57 & 520 & 3.5 & 3    \\ \cline{3-8} 
                                                & \multirow{-2}{*}{DD}                         & F & 2  & 54 & 508 & 3.6 & 2.5  \\ \cline{2-8} 
                                                &                                              & M & 12 & 56 & 165 & 2.6 & 3    \\ \cline{3-8} 
\multirow{-6}{*}{Eigsti}                        & \multirow{-2}{*}{ASD}                        & F & 4  & 55 & 145 & 2.9 & 3    \\ \hline
\rowcolor[HTML]{FFFFFF} 
\cellcolor[HTML]{FFFFFF}                        & \cellcolor[HTML]{FFFFFF}                     & M & 15 & 29 & 161 & 2.4 & 2    \\ \cline{3-8} 
\cellcolor[HTML]{FFFFFF}                        & \multirow{-2}{*}{\cellcolor[HTML]{FFFFFF}TD} & F & 11 & 34 & 233 & 2.8 & 2.3  \\ \cline{2-8} 
\cellcolor[HTML]{FFFFFF}                        &                                              & M & 10 & 61 & 120 & 2   & 2    \\ \cline{3-8} 
\multirow{-4}{*}{\cellcolor[HTML]{FFFFFF}Nadig} & \multirow{-2}{*}{ASD}                        & F & 2  & 53 & 68  & 1.6 & 1.5  \\ \hline
\end{tabular}
\caption{\label{fig:table1} ASDBank Data with participant profiles and key features.
Definitions: [1] Age in months, [2] Number of words spoken by a speaker in all the interactions, [3] Average number of words per speaker’s utterance, [4] Median number of words per speaker’s utterance.
}
\end{table*}

A number of non-invasive, early-detection technologies using artificial intelligence and machine learning [6] are being attempted to accurately and timely diagnose ASD in children [7]. For example, facial features of children during their social interactions [8] at the onset of the disease display unique and differential characteristics compared to normal children [9]. Researchers have successfully used computer vision to train machine learning models that can detect ASD using facial images of impacted children with a reasonable degree of accuracy [10]. However, one of the major drawbacks of using facial images of children is the invasion of privacy, especially for minors and the need to collect their images over prolonged periods of their childhood. Attempts have also been made to diagnose ASD using retinal image [11] and brain image [12] analysis as objective screening methods.

Unlike images, children’s speech data can be leveraged to detect language disorders [13] and ASD [14] while maintaining their privacy. Like fingerprints, every child’s linguistic patterns are unique. Studies have shown that speech patterns in ASD children change abnormally with varying levels of their voice pitch and spectral content. One of the characteristics of Autistic children is that they tend to repeat certain words and phrases over and over. These words and phrases are regular and simple, typical of the vocabulary of children in their age group. These speech abnormalities provide an excellent opportunity to use computational linguistics and machine learning [15], [16], [17], [18] to detect early and diagnose ASD with reasonable accuracy.

In this research, we are attempting to develop a number of machine learning models to detect ASD using children’s speech patterns and anomalies. We explore the effectiveness of our models using these speech patterns and data from children with Typical Development (TD) and children with ASD. Since all machine learning models depend extensively on relevant and actionable data, we spent a great deal of research time exploring, identifying, and evaluating the most appropriate data sets to test our models.

\begin{figure*}

\centering
\includegraphics[width=1.0\textwidth]{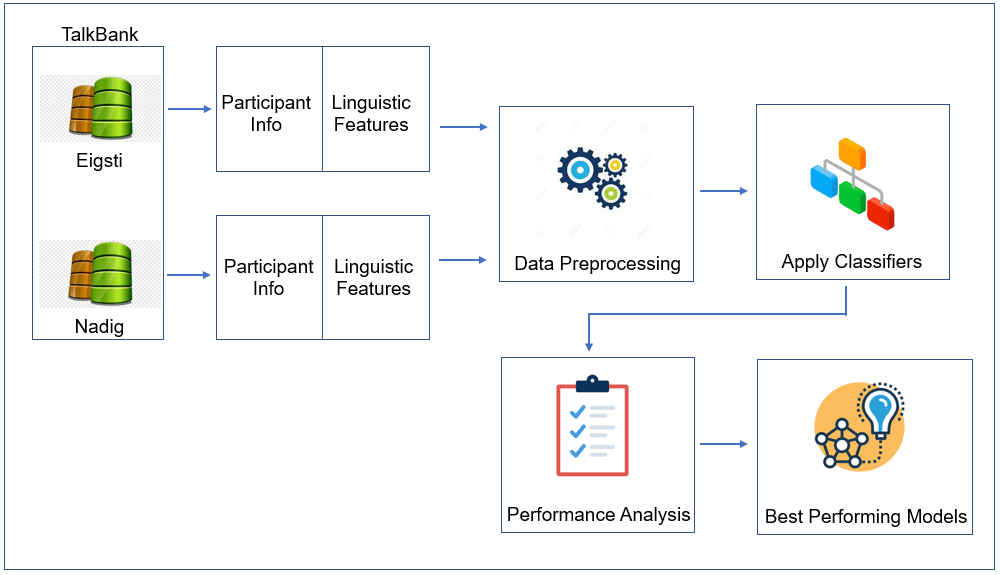}

\caption{\label{fig:figure1} Process for detecting ASD in children using computational linguistics and machine learning 
models.}
\end{figure*}

\section{Materials and Methods}

\subsection{Datasets}

A vast majority of data sets available today are pertaining to the use of written language and digital images; extensive data collection and analysis on natural spoken language, especially children, is only now beginning to gain traction.

For our research and the machine learning models we developed, we were able to use data from the TalkBank system as the primary source. The TalkBank system is a repository for spoken language data and is considered to be the world’s largest open-access and integrated database. A number of researchers across a wide variety of scientific domains that include education, medicine, computer science, linguistics, psychology, speech sciences, etc., leverage TalkBank’s language corpora [19]. Many types of spoken-language multimedia data of children and adolescents are available for access online in this system. We used the CHILDES data bank and ASDBank English database for our research to predict autism spectrum disorder since the methods and analysis used are similar in both [20], [21].

The datasets used for our analysis from Eigsti [22] consisted of children with typical development, children with non-ASD developmental delay, and children with ASD in the age group of 3-6 years. A total of 48 children were in this dataset, where each child had a 30-minute free play session. The children considered had the following ethnicities: White - 39, LatinX - 3, African-American - 6. The average age of participants in this dataset was 51 months with an average utterance of 3.08 words and a median utterance of 2.6 words.

Datasets used from Nadig, Bang [23] included 38 children and consisted of both ASD and TD children from English-speaking families. Natural Language samples were collected in this study for over a year at three different points during parent-child interactions [24]. Data from children with ASD (36-74 months) and TD children (12-57 months) were collected. Since children with ASD go through significant language delays, they were substantially older than TD children as a group.

We selected these two data sets primarily because of:

  1. the relevant age group of the data of both ASD and TD cohorts,

2. satisfactory number of participants compared to the rest of the available data and hence an adequate number of data points, and

  3. the availability of both ASD and TD child data in the corpus.

These data requirements were critical to achieving a higher quality of results from our machine learning models.

We used TalkBank’s API’s to get the data corpus from both these sources. Four different API calls were made for getting the participant details, word tokens, word utterances, and transcript information. Raw data from the two ASD TalkBank sources were cleansed and prepared for use. Linguistic tokens were identified in terms of stem words and their parts of speech and were added to the datasets as features for our machine learning models. The datasets used have a total of 61 and 52 features respectively, including participant and linguistic attributes with summary features as depicted in Table 1. Selection of the right features was critical for accurate results since speech differs in substantially unique ways vis-a-vis written words [25].

\subsection{Machine Learning Models}

\begin{table*}[ht]
\begin{center}
\begin{tabular}{||c c c c c||} 
\hline
Dataset & Participants & TD & ASD & DD \\
 \hline\hline
 Nadig & 38 & 26 & 12 & None \\ 
 \hline
 Eigsti & 48 & 16 & 16 & 16  \\
 \hline
\end{tabular}
\end{center}
\caption{\label{fig:table2} Data profiles of the ASD datasets from TalkBank}
\end{table*}

To perform our experiments/analyses, we used a number of machine learning models. Our basis for choosing these models was based on data processing, overfitting, tuning, and interpretability. 
We used cloud-based platforms like Google Colab, leveraging data from ASD TalkBank as laid out in the data section, to run our algorithms based on the process as shown in Figure 1.
Data pre-processing was performed by cleansing and adding mean values to features with any missing data. Also, categorical features were encoded by corresponding integer values. At the outset, one of the issues with the data is the class imbalance that was addressed by using SMOTE, which synthesizes new minority classes or re-weights the data. We used SMOTE by synthesizing new minority classes so that we could have a balanced data set. SMOTE was set to random state of 0 to ensure the same split for every run. The default sampling strategy was set to 'auto' indicating that we re-sampled all classes except the majority class. 

A total of 5 classifiers were applied to these transformed datasets and were selected for our final research and analysis:\\

\textbf{a. Support Vector Machine (SVM)} \\SVM is an algorithm that is used to classify both linear and nonlinear data. Typically SVM performs well with high-dimensional datasets using non-linear mapping. It explores an optimal separating hyper plane (decision boundary) of one class to another. When a radial basis function is used as a kernel, SVM automatically determines centers, weights, and thresholds and minimizes an upper bound of expected test error [26]. \\

\textbf{b. Random Forest (RF)}\\Significant improvements in classification accuracy have resulted from growing an ensemble of trees and letting them vote for the most popular class. A random forest fits a number of decision tree classifiers by sub-sampling dataset to improve the predictive accuracy of the model and also control for over-fitting [27].\\

\textbf{c. Naive Bayes (NB)} \\Naive Bayes was originally developed based on the principles of Bayes' theorem. This supervised machine learning algorithm simplifies the learning process by assuming that features are independent given the class variable. In practice, Naive Bayes often competes well with more sophisticated classifiers [28].  \\

\textbf{d. Logistic Regression (LR)}\\ Logistic regression is a predictive algorithm that applies the principles and techniques of linear regression to solve classification problems. Models based on logistic regression output a discrete value - 0 or 1, Yes or No, etc., using independent variables to predict the dependent variable [29]. \\

\textbf{e. K-Nearest Neighbor (KNN)}\\ KNN classifier is used to predict the label of a new data point by looking at its proximity to other labeled observations and assigning it to the class of the most similar labeled. The KNN function uses a Euclidean distance method to estimate the distances between the unlabeled data point and all the labeled observations that are closest to it. The maximum votes it gets for the type of labeled data points closest to it will determine how the new data point will be classified [30].

\section{Experimental Results}

\begin{table*}[ht]
\begin{tabular}{|l|l|c|c|c|c|c|}
\hline
\multicolumn{1}{|c|}{\textbf{Talk Bank Datasets}} &
  \multicolumn{1}{c|}{\textbf{Metric}} &
  \textbf{\begin{tabular}[c]{@{}c@{}}Logistic\\  Regression\end{tabular}} &
  \textbf{SVM} &
  \textbf{Random Forest} &
  \textbf{Naive Bayes} &
  \textbf{K-NN} \\ \hline
\multirow{6}{*}{Nadig}  & Precision     & \textbf{0.83} & 0.83 & 0.75          & 0.50          & \textbf{0.83} \\ \cline{2-7} 
                        & Recall        & \textbf{0.75} & 0.75 & 0.75          & 0.50          & \textbf{0.75} \\ \cline{2-7} 
                        & F1 (weighted) & \textbf{0.73} & 0.73 & 0.75          & 0.50          & \textbf{0.73} \\ \cline{2-7} 
                        & Accuracy      & \textbf{0.75} & 0.75 & 0.75          & 0.50          & \textbf{0.75} \\ \cline{2-7} 
                        & ROC-AUC       & \textbf{0.75} & 0.50 & 0.75          & 0.50          & \textbf{0.75} \\ \cline{2-7} 
                        & \multicolumn{6}{l|}{}                                                                \\ \hline
\multirow{6}{*}{Eigsti} & Precision     & 0.86          & 0.85 & \textbf{0.64} & 0.64          & 0.04          \\ \cline{2-7} 
                        & Recall        & 0.60          & 0.40 & \textbf{0.80} & 0.80          & 0.40          \\ \cline{2-7} 
                        & F1 (weighted) & 0.63          & 0.40 & \textbf{0.71} & 0.71          & 0.06          \\ \cline{2-7} 
                        & Accuracy      & 0.60          & 0.40 & \textbf{0.80} & 0.80          & 0.20          \\ \cline{2-7} 
                        & ROC-AUC       & 0.50          & 0.40 & \textbf{0.75} & 0.50          & 0.50          \\ \cline{2-7} 
                        & \multicolumn{6}{c|}{}                                                                \\ \hline
\multirow{5}{*}{Eigsti(Multi)} &
  Precision &
  \multirow{5}{*}{\begin{tabular}[c]{@{}c@{}}Low \\ Score\end{tabular}} &
  \multirow{5}{*}{\begin{tabular}[c]{@{}c@{}}Low \\ Score\end{tabular}} &
  0.40 &
  \textbf{0.71} &
  \multirow{5}{*}{\begin{tabular}[c]{@{}c@{}}Low\\ Score\end{tabular}} \\ \cline{2-2} \cline{5-6}
                        & Recall        &               &      & 0.30          & \textbf{0.50} &               \\ \cline{2-2} \cline{5-6}
                        & F1 (weighted) &               &      & 0.50          & \textbf{0.50} &               \\ \cline{2-2} \cline{5-6}
                        & Accuracy      &               &      & 0.50          & \textbf{0.60} &               \\ \cline{2-2} \cline{5-6}
                        & ROC-AUC       &               &      & 0.50          & \textbf{0.60} &               \\ \hline
\end{tabular}
\caption{\label{fig:table3}Results from various models for different performance metrics. All performance metrics are evaluated over different ASD datasets from TalkBank. Low score values indicated are less than 0.40.
}
\end{table*}
 
\subsection{Quantitative Evaluation}

To evaluate our models, we used datasets primarily from ASD TalkBank. The Nadig dataset had a total of 38 participants who were primarily children with ASD and TD children. The Eigsti data is more comprehensive and includes data of children with Delayed Development (DD). This is important to consider, since in many instances, ASD and DD characteristics overlap leading to misdiagnosis. A summary of the number of participants and their characteristics is provided in Table 2.
In Machine Learning, classifier performance measures are used to assess how well the algorithms perform for a given dataset. These measures provide us the framework for evaluating the strength and limitations of the model predictions. 
We have used 5 different metrics to evaluate the performance of our models and establish their effectiveness, as depicted in Table 3. Classifiers that showed accuracy below 40\% were omitted from the results in our report. \\
\\
\textbf{True Positives (TP):} Total number of times the model correctly predicts positive outcomes.\\
\textbf{False Positives (FP):} Total number of times the model incorrectly predicts as positive.  \\
\textbf{True Negatives (TN):} Total Number of times the model correctly predicts negative outcomes.  \\
\textbf{False Negatives (FN):} Total Number of times the model incorrectly predicts as negative.\\

\textbf{Precision:} [31] is the ratio that measures the effectiveness of the model in predicting positive labels out of all positive predictions made [31].
\[Precision = \frac{TP}{TP + FP}\] \\
\textbf{Recall:} Measures the proportion of actual positives that were identified correctly. The model correctly predicts the positives out of actual positives [31].
\[Recall = \frac{TP}{TP + FN} \]  \\
\textbf{Accuracy:} Measures the total number of model outcomes that were predicted correctly and how often we can expect our machine learning model will correctly predict an outcome out of the total number of times it made predictions [31].
\[Accuracy = \frac{TP + TN}{ TP + FN + TN + FP}\]\\
\textbf{F1 Score:} Is calculated based on both Precision and Recall and measures the accuracy of the model. This metric provides a more accurate measure of the incorrectly classified outcomes. 
\[F1 Score = \frac{2* Precision Score * Recall Score}{Precision Score + Recall Score}\] \\
\textbf{Receiver operating characteristic (ROC):}\\ROC curve helps in visualizing how the models are performing by plotting a graph with the False Positive and True Positive Rates on the x-axis and y-axis, respectively. The area under the ROC curve (AUC) determines how well the model has performed. The higher the area under the curve is, the better the performance [32], [33].
Results from our experiments for each of the classifiers against the 5 performance metrics are captured in Table 3. Of particular significance is the consistency of results achieved for each of these models across the various performance measures.\\
 
\begin{figure*}[ht]
\begin{minipage}[c]{0.5\linewidth}
\includegraphics[width=\linewidth]{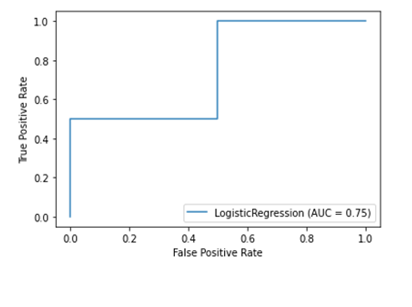}
\caption{AUC-Logistic Regression(Nadig Dataset)}
\end{minipage}
\hfill
\begin{minipage}[c]{0.5\linewidth}
\includegraphics[width=\linewidth]{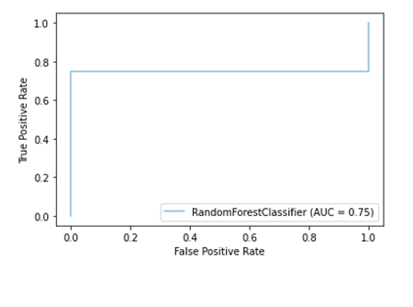}
\caption{AUC-Random Forest(Eigsti Dataset)}
\end{minipage}%
\end{figure*}

\textbf{Nadig dataset} : After running the experiment on 5 different classifiers, we identified logistic regression and KNN to have very similar results. We did not consider SVM due to the low ROC-AUC result of 0.50 since we do not want to classify TD children as ASD. Model accuracy was 0.75, and the F-1 score was 0.73. For the logistic regression model, hyper-parameter tuning was performed by increasing  the max iterations from the default value of 100 to 750, for the solver to converge. Similarly, for our KNN model, the number of neighbors was set to 5 and all points in the neighbourhood were set to have equal weights.
Having a high recall for the model is important since we want to ensure that ASD kids are identified accurately and at the same time not mis-classify ASD children as TD. Recall was found to be 0.75. The Precision result was 0.83. We also looked at the ROC-AUC to understand what proportion of the typical children got incorrectly classified as ASD. Figured 5. shows 0.75 AUC indicating reasonable accuracy of our models.

\textbf{Eigsti dataset} :After running the experiment on 5 different classifiers, we identified random forest to have the best results. Model accuracy was 0.80, and the F-1 score was 0.71. Recall was found to be 0.80, and the Precision result was 0.64. We also looked at the ROC-AUC to understand what proportion of the typical children got incorrectly classified as ASD. Figure 6. shows 0.75 AUC indicating reasonable accuracy of our models. Hyper-parameter tuning for the random forest model was performed by increasing the n estimator, which indicates the number of trees in the forest, from the default value of 100 to 200, and the max depth of the tree was set to 2.

\textbf{Eigsti (multi) dataset} :In this dataset, children with delayed language development were also considered along with ASD and TD children. Naive Bayes was found to be the most accurate model for this dataset with a Precision of 0.71, Accuracy of 0.60, and ROC-AUC of 0.60. Three of the five classifiers resulted in low scores of less than 0.40 for Accuracy and hence were not considered. Gaussian Naive Bayes classifier was used with

\subsection{Feature Importance}
 
Some of the critical features in the datasets included data about the parent’s education and linguistic abilities [34]. We used the Recursive Feature Elimination (RFE) [35] method to identify the most relevant attributes in predicting the target variable for the various models. RFE is a feature selection algorithm that iteratively finds the most relevant features from the parameters of a learnt ML model.

The following rank in the top four of the most important features that are in common in both Eigsti and Nadig datasets. \\
a. \textbf{MLU} is the Mean Length Utterance, and it measures the number of morphemes to the number of utterances [36] of the child. A morpheme is the smallest meaningful lexical item in a language as in, e.g. in, come, -ing, forming incoming [37].\\
b. \textbf{MLT} ratio is the mean length of a turn. A turn is defined as the number of utterances in a sequence spoken by the child. MLT ratio is the ratio between the mean length of the turns of the child and the mother/investigator.  A ratio of 1.0 indicates that both participants have contributed equally to the discussion.  A ratio greater than 1.0 indicates that the mother/investigator has dominated the conversation. \\
c. \textbf{Parts of speech (POS)} including percentage of pronouns, conjunctions, and negations spoken by the child. \\
d. \textbf{Number of words} spoken by the child.

\section{Conclusion}

Our primary goal was to accurately identify children with ASD early, using their linguistic data and machine learning algorithms to achieve higher orders of predictability. We focused our research on the most appropriate and relevant datasets to enhance the accuracy of our predictive models.  We incorporated a total of 57 features in the first and 61 for the second dataset to improve the efficacy of our models against 5 performance measures. We confirmed accuracy of greater than 0.75 for both datasets that make these non-invasive methods of diagnosing ASD more promising.

For future directions, it would be significant to continue to improve the accuracy of the models by adding new linguistic features as well as evaluating various methods, algorithms, and classifiers. Of particular interest are the low scores obtained for 3 of the models for the Eigsti (multi) dataset. Achieving higher orders of prediction accuracy for this dataset will be of special consequence since this is the only dataset with DD data included and is also one of the reasons for the misdiagnosis of ASD. One of the major limitations was the availability of data pertaining to children in the age group of only 3-6 years. To overcome this, we need larger datasets that capture speech in diverse environments and settings to achieve higher orders of timely predictions across a much broader age group of children.

Our goal is to continue the research by performing additional experiments to achieve a reliable and accurate non-invasive method of early ASD detection in children using computational linguistics. Future systems for detection could possibly be an interaction of children with chatbots [39] or robotic assistants [40] enabled with conversational AI for automated screening and prediction and recommending appropriate treatment plans.
\\
\bibliographystyle{unsrt}
\bibliography{research.bib}

\begin{thebibliography}{10}

\bibitem{CDC1}
CDC.
\newblock Data and statistics on autism spectrum disorder,.
\newblock {\em Center for Disease Control}, 2020.

\bibitem{MMWR2}
Matthew et~al Maenner.
\newblock Prevalence of autism spectrum disorder among children aged 8 years
  — autism and developmental disabilities monitoring network.
\newblock {\em CDC}, 2016.

\bibitem{JAMA3}
Geraldine Dawson and Guillermo Sapiro.
\newblock Potential for digital behavioral measurement tools to transform the
  detection and diagnosis of autism spectrum disorder.
\newblock {\em JAMA Pediatrics}, 173(4):305--306, 2019.

\bibitem{PRBM4}
Jennifer Elder, Consuelo Kreider, Susan Brasher, and Margaret Ansell.
\newblock Clinical impact of early diagnosis of autism on the prognosis and
  parent-child relationships.
\newblock {\em Psychology Research and Behavior Management}, Volume
  10:283--292, 08 2017.

\bibitem{TGH5}
Stephen~Bauer MD.
\newblock Asperger syndrome.
\newblock {\em The Developmental Unit The Genesee Hospital Rochester, New
  York}, 13(1), 1996.

\bibitem{IJFGCN6}
Mamata~V. Lohar and Suvarna~S. Chorage.
\newblock Machine learning-based models for early stage detection of autism
  spectrum disorders.
\newblock {\em International Journal of Future Generation Communication and
  Networking}, 13(1):426--438, 2020.

\bibitem{IEEE7}
Tania Akter, Md~Satu, Md.~Imran Khan, Mohammad Ali, Shahadat Uddin, Pietro Lio,
  Julian Quinn, and Mohammad~Ali Moni.
\newblock Machine learning-based models for early stage detection of autism
  spectrum disorders.
\newblock {\em IEEE Access}, PP, 11 2019.

\bibitem{JND8}
Blythe Corbett, Cassandra Newsom, Alexandra Key, Lydia Qualls, and E.~Kale
  Edmiston.
\newblock Examining the relationship between face processing and social
  interaction behavior in children with and without autism spectrum disorder.
\newblock {\em Journal of Neurodevelopmental Disorders}, 6:35, 08 2014.

\bibitem{MA9}
Charline Grossard, Arnaud Dapogny, David Cohen, Sacha Bernheim, Estelle
  Juillet, Fanny Hamel, Stephanie Hun, Jérémy Bourgeois, Hugues Pellerin,
  Sylvie Serret, Kevin Bailly, and Laurence Chaby.
\newblock Children with autism spectrum disorder produce more ambiguous and
  less socially meaningful facial expressions: an experimental study using
  random forest classifiers.
\newblock {\em Molecular Autism}, 11, 01 2020.

\bibitem{IJAR10}
Jahanara Shaik and Shobana Padmanabhan.
\newblock Detecting autism from facial images.
\newblock {\em International Journal of Advanced Research}, 03 2021.

\bibitem{LAN11}
Maria Laia, Jack Lee, Sally Chiu, Jessie Charm, Wing~Yee So, Fung~Ping Yuen,
  Chloe Kwok, Jasmine Tsoi, Yuqi Lin, and Benny Zee.
\newblock A machine learning approach for retinal images analysis as an
  objective screening method for children with autism spectrum disorder.
\newblock {\em EClinical Medicine, The Lancet}, 28, 2020.

\bibitem{RIN12}
Hidir~Selcuk Nogay and Hojjat Adeli.
\newblock Machine learning (ml) for the diagnosis of autism spectrum disorder
  (asd) using brain imaging.
\newblock {\em Reviews in the Neurosciences}, 31(8):825--841, 2020.

\bibitem{ASLHA13}
B~MacWhinney, A~Holland, Ratner~N. B, M~Forbes, D~Fromm, L~Togher, and
  M~Bourgeois.
\newblock Language banking for language disorders.
\newblock {\em American Speech-Language-Hearing Association Convention}, 2014.

\bibitem{BRM14}
Brian Macwhinney.
\newblock Understanding spoken language through talkbank.
\newblock {\em Behavior Research Methods}, 51, 12 2018.

\bibitem{LIN15}
Hiroki Tanaka, Sakriani Sakti, Graham Neubig, Tomoki Toda, and Satoshi
  Nakamura.
\newblock Linguistic and acoustic features for automatic identification of
  autism spectrum disorders in children's narrative.
\newblock {\em Workshop on Computational Linguistics and Clinical Psychology:
  From Linguistic Signal to Clinical Reality}, 2014.

\bibitem{UNCCH16}
Sweta Karlekar, Tong Niu, and Mohit Bansal.
\newblock Detecting linguistic characteristics of alzheimer's dementia by
  interpreting neural models.
\newblock {\em University of North Carolina at Chapel Hill}, 04 2018.

\bibitem{BEA17}
Khairun-nisa Hassanali and Yang Liu.
\newblock Measuring language development in early childhood education: a case
  study of grammar checking in child language transcripts.
\newblock {\em Proceedings of the 6th Workshop on Innovative Use of NLP for
  Building Educational Applications}, pages 87--95, 06 2011.

\bibitem{FL18}
Angela~Xiaoxue He, Rhiannon Luyster, Sung Hong, and Sudha Arunachalam.
\newblock Personal pronoun usage in maternal input to infants at high vs. low
  risk for autism spectrum disorder.
\newblock {\em First Language}, 38:014272371878263, 06 2018.

\bibitem{CMU19}
Brian MacWhinney.
\newblock The childes project: Tools for analyzing talk.
\newblock {\em Carnegie Mellon University}, 2021.

\bibitem{Childes20}
Alessandro Sanchez, Stephan Meylan, Mika Braginsky, Kyle Macdonald, Daniel
  Yurovsky, and Michael Frank.
\newblock childes-db: a flexible and reproducible interface to the child
  language data exchange system.
\newblock {\em Carnegie Mellon University}, 04 2018.

\bibitem{TB21}
Brian Macwhinney, Steven Bird, Christopher Cieri, and Craig Martell.
\newblock Talkbank: Building an open unified multimodal database of
  communicative interaction.
\newblock {\em Carnegie Mellon University}, 01 2004.

\bibitem{JADD22}
Inge-Marie Eigsti, Loisa Bennetto, and Mamta Dadlani.
\newblock Beyond pragmatics: Morphosyntactic development in autism.
\newblock {\em Journal of autism and developmental disorders}, 37:1007--23, 08
  2007.

\bibitem{ISAR23}
Janet Bang and Aparna Nadig.
\newblock Learning language in autism: Maternal linguistic input contributes to
  later vocabulary.
\newblock {\em Autism research : official journal of the International Society
  for Autism Research}, 8, 03 2015.

\bibitem{JCD24}
Aparna Nadig and Anjali Mulligan.
\newblock Intact non-word repetition and similar error patterns in
  language-matched children with autism spectrum disorders: A pilot study.
\newblock {\em Journal of Communications Disorders}, pages 13--21, 03 2017.

\bibitem{LSA25}
Paul De~Palma, Leon Garcia-Camargo, Jeb Kilfoyle, Mark Vandam, and Joseph
  Stover.
\newblock Speech tested for zipfian fit using rigorous statistical techniques.
\newblock {\em Proceedings of the Linguistic Society of America}, 6:394, 03
  2021.

\bibitem{MLA26}
Theodoros Evgeniou and Massimiliano Pontil.
\newblock Support vector machines: Theory and applications.
\newblock {\em Machine Learning and Its Applications, Advanced Lectures},
  2049:249--257, 01 2001.

\bibitem{UCB27}
Leo Brieman.
\newblock Random forests.
\newblock {\em Statistics Department, University of California, Berkeley},
  2001.

\bibitem{IJCAI28}
Irina Rish.
\newblock An empirical study of the naïve bayes classifier.
\newblock {\em IJCAI 2001 Work Empir Methods Artif Intell}, 3, 01 2001.

\bibitem{EML29}
Claude Sammut and Geoffrey~I. Webb.
\newblock Logistic regression".
\newblock {\em Encyclopedia of Machine Learning and Data Mining}, pages
  780--781, 2017.

\bibitem{ATM30}
Zhongheng Zhang.
\newblock Introduction to machine learning: k-nearest neighbors.
\newblock {\em Annals of Translation Medicine}, 4(11):218, 2016.

\bibitem{IJMLT31}
David M.~W. Powers.
\newblock Evaluation: From precision, recall and f-measure to roc,
  informedness, markedness and correlation.
\newblock {\em International Journal of Machine Learning Technology},
  2(1):37--63, 2011.

\bibitem{RSNA32}
David Vining and Gregory Gladish.
\newblock Receiver operating characteristic curves: a basic understanding.
\newblock {\em Radiographics : a review publication of the Radiological Society
  of North America, Inc}, 12:1147--54, 12 1992.

\bibitem{FLR33}
Elizabeth Krupinski.
\newblock Receiver operating characteristic (roc) analysis.
\newblock {\em Frontline Learning Research}, 5:31--42, 07 2017.

\bibitem{AI34}
Avrim~L. Blum and Pat Langley.
\newblock Selection of relevant features and examples in machine learning.
\newblock {\em Artificial Intelligence}, 97(1):245--271, 1997.

\bibitem{IJET35}
Puneet Misra and Arun Singh.
\newblock Improving the classification accuracy using recursive feature
  elimination with cross-validation.
\newblock {\em International Journal on Emerging Technologies}, 11:659--665, 05
  2020.

\bibitem{SIIPR36}
Apni Ranti, Tadris Bahasa, and Karya~Ilmiah Inggris.
\newblock Mean length utterance of children morphological development.
\newblock {\em The 1st National Conference on English Language Teaching
  (NACELT): Applied Linguistics, General Linguistics, and Literature}, 2015.

\bibitem{SAJCD37}
Helena Oosthuizen and Frenette Southwood.
\newblock Methodological issues in the calculation of mean length of utterance.
\newblock {\em South African Journal of Communication Disorders}, 56(1), 2009.

\bibitem{BIU38}
Dafna Yitzhaki.
\newblock Language profiles of children with sli using the childes system.
\newblock {\em Bar Ilan University, Israel}, 2002.

\bibitem{AMIA39}
Yue You and Xinning Gui.
\newblock Self-diagnosis through ai-enabled chatbot-based symptom checkers:
  User experiences and design considerations.
\newblock {\em AMIA Annual Symposium Proceedings}, 2021.

\bibitem{JMIR40}
Amelia Fiske, Peter Henningsen, and Alena Buyx.
\newblock Your robot therapist will see you now: Ethical implications of
  embodied artificial intelligence in psychiatry, psychology, and
  psychotherapy.
\newblock {\em Journal of Medical Internet Research}, 21, 2019.

\end{thebibliography}
\cite {CDC1}
\cite {MMWR2}
\cite {JAMA3}
\cite {JAMA3}
\cite {PRBM4}
\cite {TGH5}
\cite {IJFGCN6}
\cite {IEEE7}
\cite {JND8}
\cite {MA9}
\cite {IJAR10}
\cite {LAN11}
\cite {RIN12}
\cite {ASLHA13}
\cite {BRM14}
\cite {LIN15}
\cite {UNCCH16}
\cite {BEA17}
\cite {FL18}
\cite {CMU19}
\cite {Childes20}
\cite {TB21}
\cite {JADD22}
\cite {ISAR23}
\cite {JCD24}
\cite {LSA25}
\cite {MLA26}
\cite {UCB27}
\cite {IJCAI28}
\cite {EML29}
\cite {ATM30}
\cite {IJMLT31}
\cite {RSNA32}
\cite {FLR33}
\cite {AI34}
\cite {IJET35}
\cite {SIIPR36}
\cite {SAJCD37}
\cite {BIU38}
\cite {AMIA39}
\cite {JMIR40}
\end{sloppypar}
\end{document}